\documentclass[11pt]{article}

\usepackage[final]{acl}

\usepackage{times}
\usepackage{latexsym}
\usepackage[inkscapelatex=false]{svg}

\usepackage[T1]{fontenc}

\usepackage[utf8]{inputenc}

\usepackage{microtype}

\usepackage{inconsolata}

\usepackage{graphicx}
\usepackage{booktabs}

\usepackage{hyperref}
\usepackage{url}
\usepackage{booktabs}
\usepackage{booktabs} 
\usepackage{multirow} 
\usepackage{graphicx} 
\usepackage{tabularx}
\usepackage[most]{tcolorbox}
\usepackage{ragged2e}
\usepackage[table]{xcolor}

 \usepackage{amsmath}
 \usepackage[capitalise]{cleveref}

\usepackage{lineno}
\usepackage{xspace}
\usepackage{xcolor}

\definecolor{darkblue}{rgb}{0, 0, 0.5}
\hypersetup{colorlinks=true, citecolor=darkblue, linkcolor=darkblue, urlcolor=darkblue}

\newcommand{\myparagraph}[1]{\noindent\textbf{#1\hspace{0.2em}}}

\usepackage[most]{tcolorbox}
\usepackage{ragged2e}
\usepackage{xcolor}

\newtcolorbox{appendixpromptbox}[1]{
enhanced,
breakable,
width=\linewidth,
colback=white,
colframe=black!45,
boxrule=0.45pt,
arc=2mm,
outer arc=2mm,
left=5pt,
right=5pt,
top=6pt,
bottom=6pt,
title={#1},
colbacktitle=black!75,
coltitle=white,
fonttitle=\bfseries\small,
before upper={
  \setlength{\parindent}{0pt}
  \setlength{\parskip}{0.45em}
  \raggedright
},
boxed title style={
boxrule=0pt,
arc=2mm,
outer arc=2mm,
left=5pt,
right=5pt,
top=3pt,
bottom=3pt
}
}

\newcommand{\method}[0]{\textsc{PragReST}\xspace}
\newcommand{\methodlong}[0]{\underline{Prag}matic \underline{Re}asoning via \underline{S}elf-\underline{T}raining \xspace}

\newcommand{\imp}[0]{\textsc{IMP-SFT}\xspace}
\newcommand{\deep}[0]{\textsc{Deep-layer-DPO}\xspace}

\newcommand{\pragmega}[0]{\textsc{PragMega}\xspace}
\newcommand{\ludwig}[0]{\textsc{Ludwig}\xspace}
\newcommand{\metoqa}[0]{\textsc{MetoQA}\xspace}
\newcommand{\altprag}[0]{\textsc{AltPrag}\xspace}

\title{\method{}: Self-Reinforcing Counterfactual Reasoning for \\ Pragmatic Language Understanding}

\author{
Jihyung Park\thanks{Equal contribution.} \quad Minchao Huang\footnotemark[1] \quad Leqi Liu \quad Elias Stengel-Eskin \\
The University of Texas at Austin
}

\begin{document}

\makeatletter
\protected\def\EMNLPnotice{%
  \begingroup
  \renewcommand{\thefootnote}{}%
  \footnotetext[0]{EMNLP Camera-Ready}%
  \endgroup
}
\gdef\@thanks{\EMNLPnotice}
\makeatother

\maketitle

\begin{abstract}
Natural language understanding often depends on meanings that are implied rather than explicitly stated, requiring pragmatic reasoning. 
Despite strong performance on math and logical reasoning, large language models (LLMs) still trail humans on pragmatic inference, often defaulting to literal interpretations.
To improve LLM pragmatic reasoning, we introduce \method{}, a self-supervised framework that constructs pragmatic QA data, generates counterfactual reasoning traces, and trains models to internalize them through supervised fine-tuning and reinforcement learning, without human-labeled training data or distillation from a stronger teacher. 
Across four pragmatic benchmarks (\pragmega, \ludwig, \metoqa, and \altprag) and multiple model families, including Qwen3, Gemma, and GPT-OSS, \method{} consistently improves over backbone models, task-specific pragmatic tuning baselines, and non-counterfactual variants of the same pipeline. On accuracy-based benchmarks, \method{} improves over the instruct backbone by 5.37\% and 5.50\% (absolute) for Qwen3-8B and Qwen3-14B, respectively.
Our error analysis and ablations underscore the importance of counterfactual reasoning: \method primarily reduces errors caused by failures to contrast observed utterances with plausible alternatives,
and removing counterfactual reasoning substantially reduces performance. 
Moreover, our training preserves out-of-domain performance on general-knowledge and mathematical reasoning benchmarks.\footnote{Code and models are available \href{https://huggingface.co/papers/2606.18624} {here}.}
\end{abstract}

\section{Introduction}
\begin{figure}[t]
    \centering
    \includegraphics[width=0.49\textwidth]{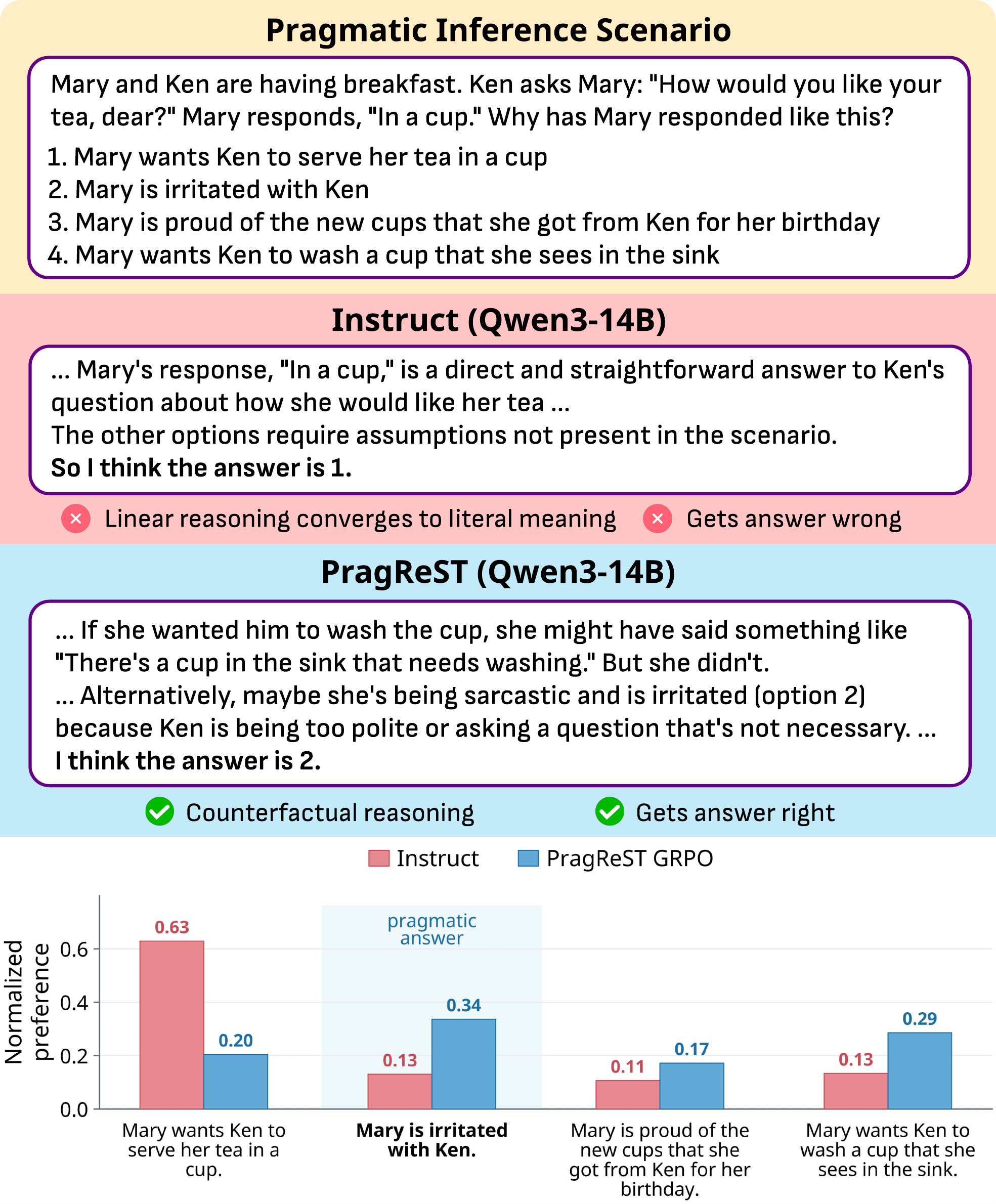}
    \vspace{-1.5em}
    \caption{
    Example of counterfactual pragmatic reasoning in \method{}. Under ordinary contextual assumptions, it recovers the intended meaning by reasoning about alternative utterances.
    Compared with the instruct model, \method{} shifts preference from the literal interpretation to the intended pragmatic interpretation.
    }
    \vspace{-1.25em}
    \label{fig:pragmatic-example}
\end{figure}
Robust language understanding requires recovering the intentions, assumptions, and implicit meanings that speakers leave unsaid \citep{grice_logic_1975}, i.e., making pragmatic inference beyond literal meaning and reasoning about shared context, speaker goals, and background information such as implicatures and presuppositions \citep{levinson_pragmatics_1983}. 
Humans can make these inferences with ease in daily conversation. Although large language models (LLMs) have improved substantially on reasoning-heavy domains such as mathematics and code \citep{shao_deepseekmath_2024, luong_reft_2024}, 
recent work suggests that they still struggle with pragmatic inference tasks \citep{ruis_goldilocks_2023, hu_fine-grained_2023, ma_pragmatics_2025, sravanthi_pub_2024}. These findings point to a persistent gap between LLMs and humans in pragmatic reasoning, suggesting that surface fluency and broad semantic competence do not, by themselves, constitute robust communicative understanding \citep{fried_pragmatics_2023}.

Cognitive science has long characterized pragmatic interpretation not as retrieval of stored meanings but as a \emph{reasoning} process. Pragmatic frameworks such as the Rational Speech Acts (RSA) \citep{frank_goodman_2012, goodman_pragmatic_2016} and Iterated Best Response (IBR) \citep{franke2009signal} cast pragmatic interpretation as recursive reasoning between speakers and listeners over communicative alternatives. 
At the center of these accounts is a pattern we refer to as \emph{counterfactual pragmatic reasoning}: a listener interprets an utterance by comparing what the speaker actually said with what they would likely have said under alternative intentions. Had the speaker intended $X$, they would likely have said $Y$; since they said $Z$ instead, $X$ becomes less plausible and an alternative meaning $W$ more plausible. This view is consistent with evidence linking counterfactual and pragmatic reasoning at the level of brain responses \citep{kulakova2016pragmatic}. 

However, translating this account into a scalable training procedure for LLMs is nontrivial for two reasons. First, unlike math and code, where Reinforcement Learning with Verifiable Rewards (RLVR) can exploit deterministic verification \citep{shao_deepseekmath_2024, luong_reft_2024}, pragmatic correctness often depends on contextual assumptions, speaker goals, and social expectations \citep{fried_pragmatics_2023, anuranjana_survey_2024}. Second, scalable pragmatic supervision is scarce: human annotation is costly, while distillation is limited by the pragmatic competence of the teacher model \citep{ma_pragmatics_2025, sravanthi_pub_2024}.

\begin{figure*}[t]
    \centering
    \includegraphics[width=0.88 \textwidth]{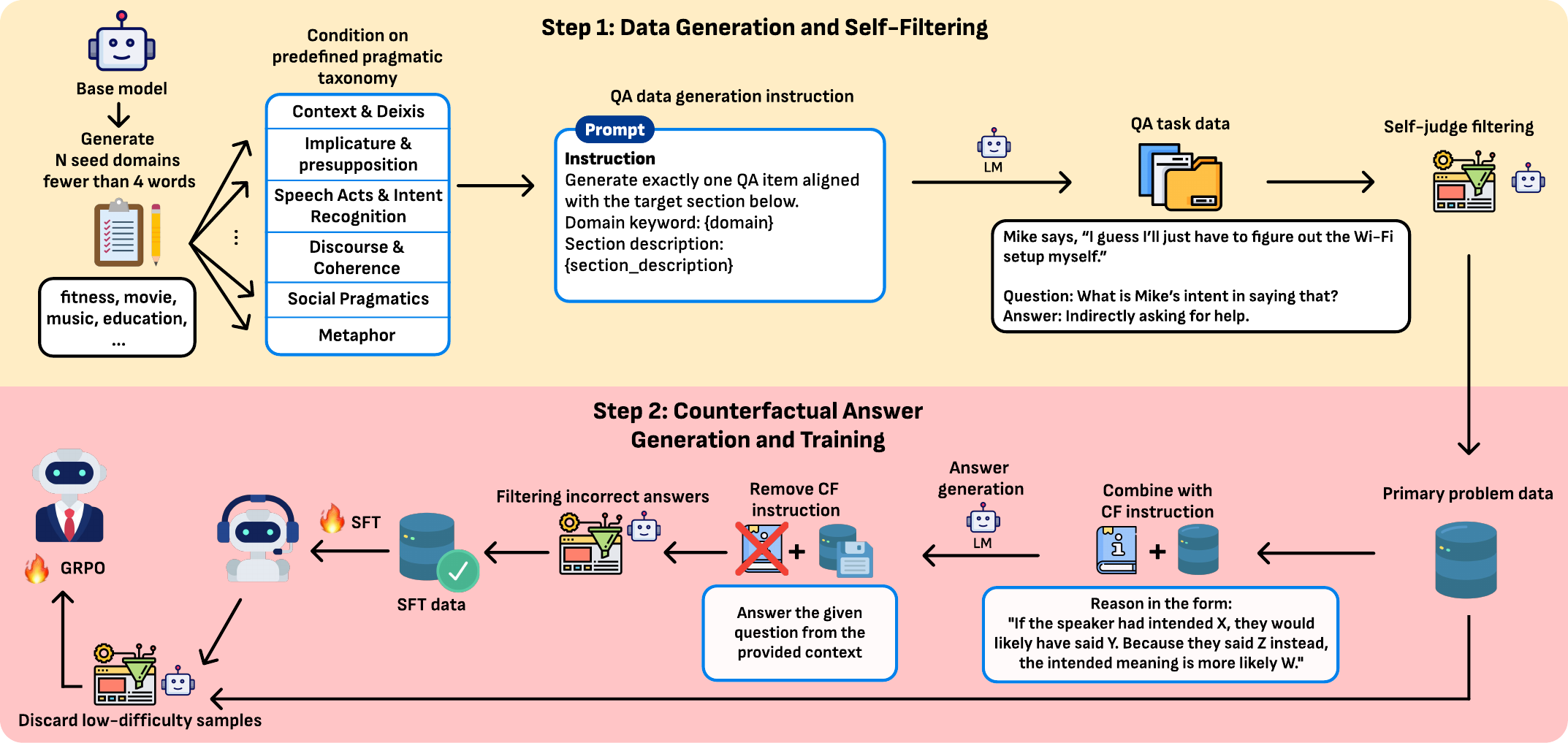}
    \caption{A high-level overview of \method. The process starts with the self-generation of domain seeds. The QA task prompt is constructed by combining a general instruction, a domain seed, and a pragmatic taxonomy description. The generated data are filtered by a self-judge. For SFT, gold answers are pre-generated with a counterfactual (CF) instruction, which is removed from the final SFT input. We also apply a second filtering step to remove examples for which the model returns an incorrect answer even with the pragmatic instruction.}
    \label{fig:pipeline}
\end{figure*}

To address these challenges, we introduce \methodlong(\method{}), a framework for learning counterfactual pragmatic reasoning from self-generated data. 
\method{} is self-reinforcing in the sense that it does not rely on human-labeled pragmatic training data or a stronger external teacher model to distill pragmatic knowledge into the policy. 
Instead, the same model is used throughout the pipeline. As illustrated in Figure~\ref{fig:pipeline}, \method{} proceeds in two stages. 
First, the model constructs a pragmatics training set and the same model audits these generated instances to remove low-quality examples. Second, \method{} turns these filtered problems into training data for teaching models counterfactual pragmatic reasoning, following two training paradigms: supervised fine-tuning (SFT) and reinforcement learning (RL).
For SFT, we first generate privileged answer traces via a prompt that explicitly encourages the model to compare the observed utterance with plausible communicative alternatives, and then train the model on these traces with the prompt removed. This teaches the model to internalize counterfactual reasoning. 
In the RL stage, we further tune the model via GRPO \citep{shao_deepseekmath_2024}, training the model on filtered problems using a self-judged correctness reward.

We evaluate \method{} on Qwen3-8B and Qwen3-14B \citep{qwen3technicalreport} and further validate its effectiveness on Gemma-4-E4B \citep{team_gemma_2026} and GPT-OSS-20B \citep{openai_gpt-oss-120b_2025}.
We evaluate across four pragmatic benchmarks: \pragmega \citep[fine-grained pragmatics QA;][]{hu_fine-grained_2023}, \ludwig \citep[implicature interpretation;][]{ruis_goldilocks_2023}, \metoqa \citep[metonymic reference resolution;][]{sravanthi_pub_2024}, and \altprag \citep[open-ended pragmatic recovery;][]
{yu_pragmatic_2026}.

To isolate the effect of counterfactual reasoning, we compare \method{} with a matched non-counterfactual variant that uses the same pipeline but replaces the counterfactual reasoning instruction with a generic pragmatic reasoning instruction. The non-counterfactual variant also improves over the instruct backbone, indicating that self-reinforcement on pragmatic data is beneficial on its own. Counterfactual supervision provides further gains: \method{} improves over the instruct backbone by an average of 5.37 and 5.50 percentage points for Qwen3-8B and Qwen3-14B, respectively, across the three accuracy-based benchmarks (\pragmega, \ludwig, \metoqa).\footnote{Unless noted otherwise, all accuracy differences are reported in terms of absolute percentages.} Notably, Qwen3-14B comes within one percentage point of the available human-performance estimates on all three benchmarks (\cref{tab:human_comparison}), and the overall improvement pattern also holds on Gemma-4-E4B, GPT-OSS-20B, and across three independent training runs (\cref{sec:Results}).

The results suggest that pragmatic self-reinforcement contributes to improved pragmatic inference, with counterfactual reasoning providing gains beyond matched non-counterfactual training. Our error analysis further shows that \method{} particularly reduces failures involving literal interpretation and missed communicative intent, where successful inference requires contrasting the observed utterance with plausible alternatives (\cref{sec:error_analysis}). Finally, these improvements do not substantially degrade out-of-domain performance on general knowledge and mathematical reasoning benchmarks, including MMLU-Pro \citep{wang_mmlu-pro_2024}, MATH-500 \citep{lightman_lets_2023, hendrycks_measuring_2021}, and AIME2025 \citep{opencompass_aime2025} (\cref{sec:ood}).

\section{Related Work}

\myparagraph{Pragmatics in Language Models.}
Pragmatic interpretation has been studied as inference over speaker intentions, contextual alternatives, and shared background assumptions.
A canonical formalization is the Rational Speech Acts (RSA) framework, which models pragmatic interpretation as probabilistic, recursive reasoning between speakers and listeners about why a speaker chose one utterance over another in a given context \citep{frank_goodman_2012, goodman_pragmatic_2016}.
This perspective has also informed computational work on pragmatically informed interpretation beyond classical reference games \citep{fried_unified_2018, vaduguru_generating_2024}.
Recent benchmarks show that even strong LLMs remain brittle on context-dependent interpretation, including implicature, presupposition, deixis, metonymic reference, and open-ended pragmatic recovery, with substantial and uneven gaps relative to humans across phenomena and settings \citep{hu_fine-grained_2023, ruis_goldilocks_2023, sravanthi_pub_2024, yu_pragmatic_2026, ma_pragmatics_2025, fried_pragmatics_2023}.
Recent training approaches improve pragmatic reasoning through task-specific preference tuning or teacher-generated thought supervision \citep{wu_rethinking_2024, sravanthi_understand_2025}.
These approaches demonstrate that pragmatic reasoning can benefit from specialized supervision, but they rely on external pragmatic data, preference annotations, or teacher-generated rationales.
In contrast, \method{} self-generates training data and uses counterfactual reasoning as the organizing principle for generating, filtering, and reinforcing reasoning traces within a self-contained training loop.

\myparagraph{Self-Improvement in Language Models.}
Prior work on LLM self-improvement trains on model-generated data, rationales, or rewards, but has focused largely on domains with deterministic tasks and verifiable rewards.
Early rationale-bootstrapping methods train models on self-generated reasoning that leads to correct answers, while later work scales this paradigm through expectation-maximization style self-training, reinforced fine-tuning, consistency-based rationale filtering, unsupervised self-training, and autonomous curriculum generation \citep{eric_zelikman_star_2022, zelikman2024quiet, singh_beyond_2023, luong_reft_2024, lee_self-training_2025, xu_genius_2025, sun_self-improvement_2025, zhao_absolute_2025}.
Synthetic-data methods such as Self-Instruct and Evol-Instruct further show that LLMs can expand their own training distributions rather than merely relabel fixed datasets \citep{wang_self-instruct_2023, xu_wizardlm_2025}.
These methods have primarily targeted math, code, or instruction-following settings in which correctness can be checked by ground-truth answers, executors, or verifier models;
\method{} extends this paradigm to open-ended pragmatics, where both the task distribution and the correctness signal must be constructed: the model generates pragmatic QA instances, filters them with a self-judge, and uses a constrained binary correctness judge rather than a general-purpose quality or preference evaluator.
This connects \method{} to LLM-based evaluators and self-rewarding systems \citep{liu_g-eval_2023, kim_prometheus_2024, yuan2024self}, while avoiding a fully general-purpose judge formulation, which can be brittle across domains \citep{zheng_judging_2023, huang_empirical_2025, raina_is_2024}.
\method{} is also related to self-distillation with privileged information and context distillation, where a teacher policy is conditioned on information unavailable to the student at inference time \citep{nguyen_avsd_2026, penaloza_privileged_2026, zhao_self-distilled_2026, ye_-policy_2026}.
In \method{}, the privileged signal is a counterfactual reasoning script used only during SFT data generation; the model is then trained on the original problem without this script, with a goal of internalizing the reasoning procedure.

\section{Methodology}

As illustrated in \cref{fig:pipeline}, \method{} consists of two stages. 
First, the model generates pragmatics-focused QA problems from domain seeds and pragmatic phenomenon descriptions, then filters the generations itself. 
Second, the model learns from this data through bootstrapped counterfactual training: in the SFT stage, we distill reasoning traces generated under a privileged counterfactual script, and in the GRPO phase we reinforce pragmatically correct answers. 
All supervision signals are produced without human-labeled pragmatic data or a stronger external model.

\subsection{Data Generation}
\label{sec:prompt-construction}
\paragraph{Prompt Construction.}
Our pipeline begins by generating a pool of short domain descriptors (e.g., Healthy Meal Prep, Modern Travel Planner).
We then sample a pragmatic taxonomy category from the list introduced by \citet{ma_pragmatics_2025}: \textit{Context and Deixis}, \textit{Implicature and Presupposition}, \textit{Speech Acts and Intent Recognition}, \textit{Discourse and Coherence}, \textit{Social Pragmatics}, and \textit{Metaphor}, as illustrated in Step 1 of \cref{fig:pipeline}.
For each domain--category pair, the model is prompted to generate an open-ended QA item consisting of a pragmatic situation, a question grounded in that situation, and a target answer. 
This differs from evaluation on fixed benchmark items. Instead of recovering an implicit meaning from an externally provided context, the model constructs the scenario, question, and intended answer together under an explicit pragmatic category description. 
The generated item can therefore be treated as a candidate locally grounded QA instance, closer to answer-aware question generation and grounded synthetic QA generation than to ordinary benchmark inference \citep{Zhangr_review_2021, radevski_synthesizing_2025}.
We expand on data generation details, including few-shot examples and prompts, in Appendix~\ref{append:data_construction}.

\paragraph{Self-Filtering.}
\label{sec:data-generation-filtering}
Using the prompts constructed in Step~1, the model generates candidate QA and dialogue instances. 
Because these candidates may still be noisy, we retain them only after filtering out examples that are malformed, ambiguous, answerable without pragmatic reasoning, or not pragmatically licensed. 
This proposal-and-filtering strategy follows prior self-generation and bootstrapping work, which uses model-generated candidates for training only after automatic filtering or correctness checks \citep{wang_self-instruct_2023, eric_zelikman_star_2022}.
For each generated instance, we use the corresponding instruct backbone as a constrained binary judge.
The judge is prompted to output either \emph{yes} or \emph{no}, indicating whether the instance meets a set of manually defined quality criteria (provided in Appendix~\ref{append:audit_filtering}).
To obtain a continuous judge signal, we derive a first-token confidence margin 
$m(q) = p(\texttt{yes} \mid q) - p(\texttt{no} \mid q)$, where $q$ is the generated QA item.
We interpret lower scores as lower-confidence generations and discard the bottom $50\%$ of the data, ranked by margin.
To give a sense of typical scale, our Qwen3-14B run starts from 1{,}000 domain seeds and yields 6{,}000 parseable QA items, of which 3{,}000 are retained after self-filtering as the primary problem data. 

\subsection{Training}
\label{sec:training}
\label{sec:sft}

We train the model in two stages. The first stage uses supervised fine-tuning (SFT) to internalize counterfactual pragmatic reasoning from generated demonstrations, and the second stage uses GRPO to reinforce pragmatically correct outcomes. We describe the main design choices below and provide full training details in Appendix~\ref{append:training}.

\paragraph{SFT with Counterfactual Bootstrapping.}
The goal of the SFT stage is to teach the model a reusable counterfactual reasoning procedure.
For each filtered training instance, we construct an augmented target-generation prompt by prepending the counterfactual reasoning script and the corresponding pragmatic section description to the original problem.
Under this augmented prompt, the model generates candidate responses consisting of a reasoning trace and a final answer; the script encourages the model to interpret an utterance as a communicative choice, compare it with plausible alternatives, and infer the speaker's intended meaning from that contrast \citep{goodman_pragmatic_2016, fried_unified_2018, vaduguru_generating_2024, tsvilodub_non-literal_2025}.
We then apply the answer-correctness judge to retain only candidate responses whose final answer is judged as being pragmatically correct with respect to the context, question, and reference answer.
At the margin threshold of $m(x,a) > 0.8$, the judge agrees with human consensus on $80.0\%$ of 100 manually reviewed examples (F1${}=0.796$; Appendix~\ref{append:judge_margin_calibration}).
For the Qwen3-14B run, 2{,}816 counterfactual responses pass this filter, yielding 2{,}759 training examples and 57 held-out synthetic validation examples.
After filtering, each accepted response is paired with the unaugmented original problem.
Both the counterfactual script and the pragmatic section description are removed from the student input.
Thus, the script acts as privileged scaffolding for constructing SFT targets, not as an inference-time prompt, and training distills the resulting counterfactual reasoning behavior into the model so that it can recover the same reasoning pattern from the original input alone.
Details on SFT data construction and hyperparameters are given in Appendix~\ref{append:sft_details}.

\paragraph{Reinforcement Learning with GRPO.}
\label{sec:grpo}

After SFT, the model has learned from counterfactual reasoning traces, but the SFT stage uses correctness only as a criterion for selecting demonstrations rather than as an objective optimized during training.
We therefore initialize from the SFT checkpoint and apply GRPO \citep{shao_deepseekmath_2024} on the filtered primary problem data. 
Before GRPO training, we apply an offline difficulty filter, following DAPO~\citep{yu_dapo_2026}. 
For each prompt, we sample $G$ rollouts from the initial SFT model and discard \emph{easy} prompts on which every rollout already passes the correctness judge, since these zero-variance groups provide no learning signal.
Note that we keep prompts for which the SFT model gets zero reward, as these may still yield some signal later from GRPO-trained checkpoints.
During GRPO training, we keep the filtered prompt set fixed but resample rollouts online as the model is updated.
For each remaining prompt, the current policy draws a fresh group of $G$ candidate responses online, which are scored using a composite reward that combines output-format compliance with pragmatic answer correctness. 
The correctness component reuses the same correctness judge and first-token margin construction defined in Appendix~\ref{append:judge_prompt}, comparing the extracted final answer against the reference answer.
This training reinforces responses that recover the intended pragmatic interpretation while regularizing the policy toward the SFT checkpoint. 
Full GRPO details and hyperparameters are provided in Appendix~\ref{append:grpo_details}.

\begin{table*}[t]
\centering
\setlength{\tabcolsep}{3pt}
\renewcommand{\arraystretch}{1.05}
\newcommand{\se}[2]{#1{\tiny\,$\pm$\,#2}}
\newcommand{\bestse}[2]{\textbf{#1}{\tiny\,$\pm$\,#2}}
\small
\begin{tabular}{lcccc}
\toprule
\textbf{Model} 
& \textbf{\pragmega} 
& \textbf{\ludwig} 
& \textbf{\metoqa}
& \textbf{\altprag} \\
\midrule
\multicolumn{5}{c}{\textit{Qwen3-8B}} \\
\midrule
Instruct 
& \se{73.37}{3.33}
& \se{80.33}{1.62}
& \se{73.52}{2.06}
& \se{7.24}{0.05} \\

\deep \citep{wu_rethinking_2024} 
& \se{67.46}{3.58}
& \se{80.00}{1.63}
& \se{72.65}{2.08}
& \se{7.24}{0.05} \\

\imp \citep{sravanthi_understand_2025}
& \se{69.82}{3.54}
& \se{82.83}{1.54}
& \se{73.09}{2.06}
& \se{7.19}{0.05} \\

\midrule

Non-counterfactual-SFT 
& \se{76.92}{3.27}
& \se{81.83}{1.57}
& \se{77.02}{1.98}
& \se{7.44}{0.05} \\

Non-counterfactual-GRPO 
& \se{77.51}{3.21}
& \bestse{83.00}{1.54}
& \se{78.12}{1.93}
& \se{7.60}{0.05} \\

\midrule
\method{}-SFT 
& \se{77.51}{3.22}
& \se{82.17}{1.55}
& \se{78.56}{1.90}
& \se{7.46}{0.05} \\

\method{}-GRPO 
& \bestse{80.47}{3.04}
& \bestse{83.00}{1.53}
& \bestse{79.87}{1.88}
& \bestse{7.62}{0.04} \\

\midrule
\multicolumn{5}{c}{\textit{Qwen3-14B}} \\
\midrule
Instruct 
& \se{81.66}{2.99}
& \se{82.67}{1.52}
& \se{71.77}{2.11}
& \se{7.78}{0.05} \\

\deep \citep{wu_rethinking_2024} 
& \se{77.51}{3.22}
& \se{83.50}{1.51}
& \se{72.87}{2.07}
& \se{7.71}{0.05} \\

\imp \citep{sravanthi_understand_2025}
& \se{81.07}{2.95}
& \se{83.17}{1.52}
& \se{72.21}{2.10}
& \se{7.53}{0.05} \\

\midrule

Non-counterfactual-SFT 
& \se{83.43}{2.88}
& \se{83.33}{1.52}
& \se{73.52}{2.06}
& \se{7.93}{0.05} \\

Non-counterfactual-GRPO 
& \se{84.62}{2.76}
& \se{84.33}{1.49}
& \se{76.15}{1.98}
& \se{8.07}{0.04} \\

\midrule
\method{}-SFT 
& \se{85.21}{2.71}
& \se{85.17}{1.45}
& \se{79.65}{1.90}
& \se{8.06}{0.04} \\

\method{}-GRPO 
& \bestse{85.80}{2.67}
& \bestse{86.50}{1.40}
& \bestse{80.31}{1.84}
& \bestse{8.14}{0.04} \\

\bottomrule

\end{tabular}%
\caption{
Performance across benchmarks and models under greedy decoding.
\pragmega, \ludwig, and \metoqa report accuracy; \altprag reports its reference-based score.
Values are point estimates with bootstrap standard errors over examples.
We bold the best point estimate for each model size and benchmark.
}
\label{tab:qwen3_pragmatic_bold}
\end{table*}

\section{Experiments and Results}

\paragraph{Benchmarks and Models.}
We evaluate our method on four benchmarks selected to directly test pragmatic interpretation: 
\pragmega, \ludwig, \metoqa, and \altprag. 
\pragmega \citep{hu_fine-grained_2023} is a QA benchmark for pragmatic language understanding spanning multiple pragmatic phenomena. 
\ludwig \citep{ruis_goldilocks_2023} evaluates implicature interpretation as a binary decision over whether a listener's response should be interpreted as yes or no. 
\metoqa \citep{sravanthi_pub_2024} evaluates metonymic reference resolution, where the model must infer the intended referent behind a contextually associated expression. 
\altprag \citep{yu_pragmatic_2026} evaluates open-ended pragmatic recovery, where the model must produce an appropriate interpretation of the implied meaning. Representative examples from each benchmark are provided in \cref{tab:benchmark_examples}.
For \pragmega, \ludwig, and \metoqa, we report accuracy;  for \altprag, we report the benchmark's reference-based evaluation score and pairwise comparisons using the original GPT-4.1 judge and scoring protocol released by the benchmark authors. 
Our primary experiments use two instruct backbone models, Qwen3-8B and Qwen3-14B; we also replicate the pipeline for two additional model families, Gemma-4-E4B and GPT-OSS-20B.
We apply the same training pipeline to each model and evaluate the resulting models under the same benchmark settings.

\paragraph{Baselines.}

We compare \method{} against four baselines. 
The first is the Instruct backbone. 
The second is \deep{}, following \citet{wu_rethinking_2024}, a layer-restricted DPO method designed for social and pragmatic inference. 
For this baseline, we follow the strongest setting reported in \citet{wu_rethinking_2024} and train on SocialIQA \citep{sap_social_2019}, a human-annotated social inference dataset. 
To adapt this baseline to reasoning models, we generate gold-guided reasoning traces by providing the correct answer during trace generation and train on the resulting reasoning-answer outputs. 
The third is \imp{}, following \citet{sravanthi_understand_2025}, which distills GPT-4o-mini-generated rationales for pragmatic understanding. 
For this baseline, we preserve the model's original reasoning behavior by masking the loss on the model-generated reasoning part, then appending the GPT-4o-mini rationale and training on the rationale paired with the correct label. 
\deep{} and \imp{} represent two recent training-based approaches to pragmatic improvement that rely on external supervision: human-annotated social inference data and teacher-rationale distillation, respectively. 
Finally, we include non-counterfactual variants of our own pipeline, which keep the same data generation, filtering, SFT, and GRPO stages as \method{} but replace the counterfactual reasoning instruction with a generic pragmatic reasoning instruction. 
This helps distinguish the effect of counterfactual reasoning from the effect of self-training and RL post-training, and provides the direct test of our central hypothesis: if counterfactual comparison is the active ingredient rather than self-training itself, then matched training without it should fail to reproduce the gains.
Counterfactual and non-counterfactual prompts are in Appendix~\ref{append:answer_generation_prompts}.

\subsection{Results}
\label{sec:Results}

Table~\ref{tab:qwen3_pragmatic_bold} shows that \method{} consistently improves pragmatic reasoning across the four evaluation benchmarks and both model sizes. 
For Qwen3-8B, \method{}-GRPO achieves the best result in nearly every benchmark, improving over the instruct backbone on \pragmega, \ludwig, \metoqa, and \altprag. 
Across the three accuracy-based benchmarks, this corresponds to an average gain of 5.37\% over the Instruct backbone; on \altprag, which uses a reference-based score, \method{}-GRPO improves from 7.24 to 7.62. 
For Qwen3-14B, \method{}-GRPO again gives the best performance across all reported benchmarks, with an average gain of 5.50\% across the three accuracy-based benchmarks and an \altprag improvement from 7.78 to 8.14. 
These results suggest that \method{} improves multiple forms of pragmatic interpretation, including fine-grained QA, implicature resolution, metonymic reference, and open-ended pragmatic recovery.

\paragraph{Comparison with Human Performance.}
To contextualize the remaining headroom on the accuracy-based pragmatic benchmarks, we compare \method-GRPO with the human-performance estimates reported or computable from the corresponding benchmark resources. We restrict this comparison to \pragmega, \ludwig, and \metoqa, since these benchmarks are evaluated with accuracy. We do not include \altprag in this table because it uses a reference-based open-ended scoring protocol rather than a directly comparable human accuracy score.
As shown in Table~\ref{tab:human_comparison}, Qwen3-14B \method-GRPO reaches performance close to the human estimates available on all three accuracy-based benchmarks. In \pragmega, the human score computed using the code of the benchmark authors over our evaluated subset is 86.37, compared to 85.80 for \method{}-GRPO. On \ludwig, \method-GRPO reaches 86.50, slightly above the reported human average of 86.2. In \metoqa, \method-GRPO reaches 80.31, close to the reported human score of 80.0. These comparisons suggest that, for the accuracy-based benchmarks, \method operates in a regime of near-human performance, which may partly explain why absolute gains over strong instruction-tuned backbones are modest on some tasks.

\begin{table}[t]
\centering
\small
\begin{tabular}{lccc}
\toprule
Benchmark & Human score & \method & Gap \\
\midrule
\pragmega & 86.37 & 85.80 & $-0.57$ \\
\ludwig & 86.20 & 86.50 & $+0.30$ \\
\metoqa & 80.00 & 80.31 & $+0.31$ \\
\bottomrule
\end{tabular}
\caption{
Comparison between Qwen3-14B \method and human-performance estimates on the three accuracy-based pragmatic benchmarks. The \pragmega human score is computed using the benchmark authors' data. The \ludwig and \metoqa human scores are those reported in the corresponding papers. 
}
\label{tab:human_comparison}
\end{table}

\paragraph{Comparison with External Baselines.}
\method{} also compares favorably against prior task-specific pragmatic tuning methods. 
For Qwen3-8B, \deep{} underperforms the instruct backbone on \pragmega, \ludwig, and \metoqa.
Meanwhile \imp{} improves over the instruct model on \ludwig, because it contains an augmented version of \ludwig in its training data, but its performance does not transfer to other benchmarks.
Overall, these comparisons underscore that \method{} improves over prior pragmatic supervision methods despite learning from a self-generated signal rather than relying on human-annotated data or a stronger teacher. 

\begin{figure}
    \centering
    \includegraphics[width=1.00\linewidth]{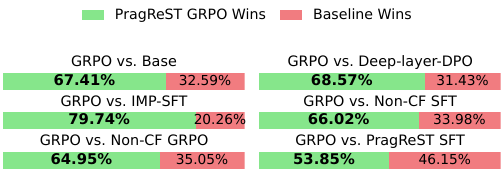}
    \caption{
    Pairwise preference on ALTPRAG: each bar compares \method-GRPO against a Qwen3-14B baseline, using GPT-4.1 as a blind pairwise judge over the two models’ generated answers.
    }
    \label{fig:altprag_pairwise}
\end{figure}

\paragraph{Preference Evaluation on \altprag.}
Because \altprag requires open-ended pragmatic recovery, we complement its reference-based score with a blind GPT-4.1 pairwise comparison over model outputs, following the benchmark's original evaluation setup.
As shown in Figure~\ref{fig:altprag_pairwise}, \method{}-GRPO is preferred over the instruct backbone in 67.41\% of decided comparisons, over the external baselines, over the non-counterfactual variants, and over \method{}-SFT in 53.85\%.
This indicates that the \altprag improvement is not only a scalar-score shift: when full interpretations are compared directly, \method{}-GRPO is more often judged to recover the intended pragmatic meaning.

\paragraph{Importance of Counterfactual Supervision.}
The non-counterfactual variants provide a controlled test of whether the gains come from self-training alone or from the counterfactual structure of the supervision. 
They use the same generated problem distribution, filtering procedure, SFT stage, and GRPO stage as \method{}, but ask the model to reason about the pragmatic meaning of the utterance in context, without providing the explicit counterfactual scaffold used in \method{}. Their weaker performance suggests that general pragmatic self-training does not match \method{}: the self-improvement loop becomes substantially more effective when the training signal teaches the model to contrast the observed utterance with plausible communicative alternatives. 
At the same time, \method{}-GRPO improves over \method{}-SFT on the primary comparison, indicating that outcome-based reinforcement adds gains beyond imitation of counterfactual traces. 
Taken together, these results suggest that the two stages play complementary roles: SFT gives the model a counterfactual reasoning procedure, while GRPO reinforces when and how to apply that procedure to recover pragmatically correct interpretations.

\begin{table}[t]
\centering
\footnotesize
\setlength{\tabcolsep}{3pt}
\renewcommand{\arraystretch}{1.05}
\newcommand{\std}[2]{#1{\tiny\,$\pm$\,#2}}
\newcommand{\beststd}[2]{\textbf{#1}{\tiny\,$\pm$\,#2}}
\resizebox{\columnwidth}{!}{%
\begin{tabular}{lcccc}
\toprule
\textbf{Model} 
& \textbf{\pragmega} 
& \textbf{\ludwig} 
& \textbf{\metoqa}
& \textbf{\altprag} \\
\midrule
\multicolumn{5}{c}{\textit{Qwen3-14B}} \\
\midrule
Instruct 
& 81.66
& 82.67
& 71.77
& 7.78 \\
\method{}-SFT 
& \std{84.62}{0.60}
& \std{84.17}{0.87}
& \beststd{80.09}{0.58}
& \std{7.83}{0.20} \\
\method{}-GRPO 
& \beststd{86.00}{0.34}
& \beststd{85.06}{1.29}
& \std{79.80}{1.08}
& \beststd{8.00}{0.12} \\
\bottomrule
\end{tabular}%
}
\caption{
Mean and standard deviation over three independent Qwen3-14B runs under greedy decoding.
The deterministic Instruct baseline is evaluated once.
}
\label{tab:qwen3_14b_variance}
\end{table}

\begin{table}[t]
\centering
\footnotesize
\setlength{\tabcolsep}{3pt}
\renewcommand{\arraystretch}{1.05}
\newcommand{\se}[2]{#1{\tiny\,$\pm$\,#2}}
\newcommand{\bestse}[2]{\textbf{#1}{\tiny\,$\pm$\,#2}}
\resizebox{\columnwidth}{!}{%
\begin{tabular}{lcccc}
\toprule
\textbf{Model} 
& \textbf{\pragmega} 
& \textbf{\ludwig} 
& \textbf{\metoqa}
& \textbf{\altprag} \\
\midrule
\multicolumn{5}{c}{\textit{Gemma-4-E4B}} \\
\midrule
Instruct 
& \se{71.60}{3.47}
& \se{82.33}{1.56}
& \se{75.05}{2.02}
& \se{7.39}{0.05} \\
\method{}-SFT 
& \se{75.15}{3.32}
& \se{82.83}{1.54}
& \se{80.96}{1.84}
& \se{7.68}{0.04} \\
\method{}-GRPO 
& \bestse{78.11}{3.18}
& \bestse{84.00}{1.50}
& \bestse{82.71}{1.77}
& \bestse{7.72}{0.04} \\
\midrule
\multicolumn{5}{c}{\textit{GPT-OSS-20B}} \\
\midrule
Instruct
& \se{68.05}{3.59}
& \se{74.33}{1.76}
& \se{74.84}{2.02}
& \se{7.41}{0.05} \\
\method{}-SFT
& \se{75.74}{3.27}
& \se{79.33}{1.66}
& \bestse{79.87}{1.85}
& \se{7.44}{0.05} \\
\method{}-GRPO
& \bestse{77.51}{3.24}
& \bestse{80.67}{1.65}
& \se{79.21}{1.90}
& \bestse{7.47}{0.05} \\
\bottomrule
\end{tabular}%
}
\caption{
Performance of \method{} on \textit{Gemma-4-E4B} and \textit{GPT-OSS-20B}.
Values are point estimates with bootstrap standard errors over examples.
}
\label{tab:gemma_gptoss_results}
\end{table}

\paragraph{Robustness and Generalization.}
\cref{tab:qwen3_14b_variance} shows that across three independent Qwen3-14B runs, both \method{}-SFT and \method{}-GRPO remain above the instruct backbone on average.
Moreover, \cref{tab:gemma_gptoss_results} shows that the same pattern holds for Gemma-4-E4B and GPT-OSS-20B, with average \method{}-GRPO gains of 5.28 and 6.72 percentage points, respectively, across the three accuracy-based benchmarks.
For GPT-OSS-20B, we use LoRA adapters and low reasoning effort due to its larger model size.
These results indicate that the gains are not specific to one training run or the Qwen3 architecture.

\section{Discussion and Analysis}

Our results in \cref{tab:qwen3_pragmatic_bold} and \cref{fig:altprag_pairwise} show that \method{} improves performance across multiple pragmatic reasoning tasks.
Matched non-counterfactual self-training also helps, but explicit counterfactual supervision yields the strongest overall results.
To better understand these trends, we analyze where \method{}'s gains stem from.
Additionally, we show that training models for pragmatic reasoning still preserves their broader reasoning and knowledge capabilities.
 
\subsection{Counterfactual Reasoning and Error Reduction}
\label{sec:error_analysis}

We test whether \method{}'s gains concentrate on errors that require comparing observed utterances with plausible alternatives.
We analyze incorrect \pragmega outputs, induce a recurring-error taxonomy, and relate changes in each error type to the amount of counterfactual reasoning observed.
Appendix~\ref{append:cf_error_analysis} provides the full construction and validation procedure.

\paragraph{Inducing an Error Taxonomy.}
Using incorrect outputs from the evaluated Qwen3 models, GPT-4.1-mini first proposes batch-level error categories and then consolidates them into five recurring types:
\emph{literal/surface bias}, \emph{missed communicative intent}, \emph{unsupported or overextended inference}, \emph{coherence-bridge error}, and \emph{figurative or humor mapping error}.
The model subsequently assigns all applicable categories to each failure case.
Definitions and prompts are provided in Appendix~\ref{append:error_taxo}.

\paragraph{Counterfactual Reasoning Score.}
In addition to error tags, each reasoning trace is automatically scored for the presence of counterfactual pragmatic reasoning. 
We prompt a GPT-4.1-mini judge to flag whether the trace considers relevant alternative utterances or interpretations, contrasts literal and intended meanings, identifies mismatches between what was said and what would have been said under a literal interpretation, and uses the speaker's communicative choice to infer intent.
A higher score means more counterfactual reasoning.

\paragraph{Validating Error-Taxonomy Annotation.}
We validate the GPT-4.1-mini error-taxonomy labels 
through a blind study of 40 error samples labeled by three humans.
Human-human agreement is 83.8\% with an average Micro Cohen's $\kappa$ of 0.628, while human-GPT agreement is 82.6\% with an average $\kappa$ of 0.614.
Appendix~\ref{append:human_valid} provides the full analysis.

\begin{figure}[t]
    \centering
    \includegraphics[width=0.9\linewidth]{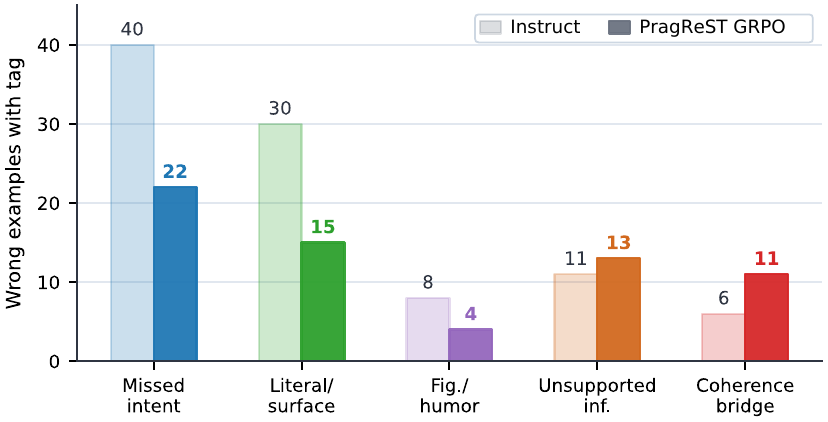}
    \caption{
    Tagged error counts before and after \method{} training.
    Bars show the number of incorrect \pragmega examples assigned to each error category for the Instruct model and \method{}.
    }
    \label{fig:error_counts}
\end{figure}

\begin{figure}[t]
    \centering
    \includegraphics[width=0.5\textwidth]{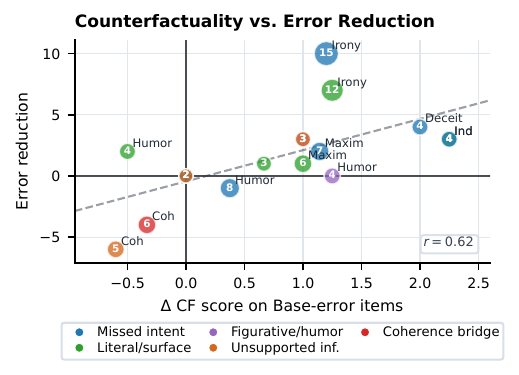}
    \caption{
    Error reduction by failure mode.
    The x-axis shows the change in mean counterfactual-reasoning score from the Instruct model to \method{}, and the y-axis shows the reduction in tagged errors. The numbers inside indicate how many errors are associated with each phenomenon covered in \pragmega.
    }
    \label{fig:error_reduction_cf}
\end{figure}

\paragraph{Error Reduction Aligns with Counterfactual Reasoning.}
As shown in \cref{fig:error_counts}, \method{} reduces the dominant counterfactual-pragmatic failure modes: missed communicative intent drops from 40 to 22, 
literal/surface bias from 30 to 15, and figurative/humor mapping from 8 to 4. 
These categories require the model to move beyond literal compatibility and infer why the speaker chose the observed utterance rather than a more direct alternative. 
The same pattern appears in \cref{fig:error_reduction_cf}: error reductions correlate with increases in counterfactual reasoning, suggesting that the gains are tied to more explicit reasoning over communicative alternatives. 
At the same time, unsupported-inference and coherence-bridge errors do not decrease, suggesting that counterfactual reasoning alone is not sufficient when the main challenge is determining whether an inferred alternative is supported by the discourse context.


\subsection{Out-of-Domain Evaluation}
\label{sec:ood}

\begin{table}[t]
    \centering
    \small

    \setlength{\tabcolsep}{4pt}
    \begin{tabular}{llcc}
    \toprule
    \textbf{Benchmark} & \textbf{Metric}
    & \textbf{Instruct} & \textbf{\method{}} \\
    \midrule
    \multicolumn{4}{c}{\textit{Qwen3-8B}} \\
    \midrule
    MMLU-Pro & Acc.
    & 71.63 & 70.72 \\
    MATH-500 & Pass@8
    & 97.40 & 97.00 \\
    AIME2025 & Pass@8
    & 80.00 & 80.00 \\
    TruthfulQA & MC2
    & 55.28 & 55.17 \\
    \midrule
    \multicolumn{4}{c}{\textit{Qwen3-14B}} \\
    \midrule
    MMLU-Pro & Acc.
    & 72.87 & 75.10 \\
    MATH-500 & Pass@8
    & 97.40 & 97.20 \\
    AIME2025 & Pass@8
    & 83.33 & 83.33 \\
    TruthfulQA & MC2
    & 55.82 & 56.01 \\
    \bottomrule
    \end{tabular}
    \caption{
    Out-of-domain evaluation for Qwen3-8B and Qwen3-14B models. MMLU-Pro accuracy is computed on a 10\% subset sampled from each subject.
    }
    \label{tab:ood_results}
\end{table}

A common concern with task-specific post-training is that improvements on the target domain may come at the cost of broader model capability. We therefore evaluate whether \method{} preserves out-of-domain performance on general knowledge, mathematical reasoning, and factual truthfulness tasks.
We compare \method{} with the instruct backbone on MMLU-Pro
\citep{wang_mmlu-pro_2024}, MATH-500
\citep{hendrycks_measuring_2021, lightman_lets_2023}, AIME2025
\citep{opencompass_aime2025}, and TruthfulQA
\citep{lin-etal-2022-truthfulqa}.
As shown in Table~\ref{tab:ood_results}, performance remains close across both model sizes, with no consistent degradation.
Thus, the pragmatic gains do not appear to systematically trade off with general knowledge, mathematical reasoning, or factual truthfulness.

\section{Conclusion}
We introduced \method{}, a self-reinforcing framework for improving pragmatic reasoning through self-generated counterfactual supervision, without human-labeled pragmatic data or stronger teacher models. 
Across pragmatic benchmarks, \method{} improves over backbone models, prior pragmatic tuning baselines, and non-counterfactual variants, with analyses showing that the gains concentrate in cases requiring comparison between what a speaker said and what they could have said under alternative intentions. 
These results suggest that reinforcement-based self-improvement can extend beyond formally verifiable domains toward socially grounded language understanding.

\section*{Limitations}

Although \method{} improves pragmatic reasoning, some limitations remain. 
First, our evaluation inherits a general limitation of pragmatics benchmarks: each item provides only a finite textual context. 
Additional background information, visual evidence, or unusual speaker assumptions could sometimes license an interpretation different from the annotated target. 
This under-determination is common in pragmatics, where even canonical examples depend on ordinary contextual assumptions.
Our experiments therefore evaluate whether models recover the intended interpretation under the assumptions encoded by existing benchmark items, not whether they resolve all possible readings under arbitrary expanded contexts.

Second, \method{} does not uniformly reduce all types of pragmatic errors. 
In our error analysis, literal/surface-bias and missed-intent errors decrease substantially, while unsupported-inference and coherence-bridge errors are not consistently reduced. 
These remaining errors suggest that generating plausible communicative alternatives is not sufficient on its own: the model must also determine whether those alternatives are supported by the specific discourse context. 
Future work could therefore incorporate stronger evidence-checking mechanisms when constructing or using counterfactual alternatives.

Finally, our evaluation is limited to English-language pragmatic benchmarks. 
This follows the available benchmark setting and allows controlled comparison with prior work, but pragmatic interpretation is strongly shaped by language, culture, social norms, and conversational conventions \citep{fried_pragmatics_2023, ma_pragmatics_2025}. 
As a result, our findings do not establish that the same counterfactual training procedure transfers to multilingual or culturally variable pragmatic settings. 
Extending \method{} to non-English and cross-cultural pragmatics is an important direction for future work.

\section*{Acknowledgments}
We would like to thank Jessy Li for her helpful feedback, and Ananya Sahu for providing annotations.

\bibliography{colm2026_conference}

\newpage
\appendix

\section{Data Construction}
\label{append:data_construction}

We construct the primary pragmatic QA data in two stages. 
First, we generate short-answer pragmatic QA instances from domain seeds and pragmatic section descriptions. 
Second, we audit the generated instances with a binary self-judging prompt and retain high-quality examples for training. 
The resulting filtered set is used as the primary problem data for both SFT target generation and GRPO training.

\subsection{Pragmatic Sections}
\label{append:pragmatic_sections}

For data generation, each example is conditioned on one pragmatic section. 
A section description is a short natural-language definition of the type of pragmatic inference the generated question should require. 
Definitions of the six sections are given in \cref{tab:pragmatic_sections}.

\begin{table*}[t]
\centering
\small
\setlength{\tabcolsep}{5pt}
\begin{tabular}{p{0.23\linewidth}p{0.70\linewidth}}
\toprule
\textbf{Section} & \textbf{Description} \\
\midrule
Context and Deixis 
& Interpret meaning that depends on context by identifying the intended referent of underspecified or deictic expressions from the dialogue and current perspective. \\

Implicature and Presupposition 
& Recover non-explicit meaning by inferring what is implied under cooperative reasoning, and by identifying background assumptions required for an utterance to be interpretable. \\

Speech Acts and Intent Recognition 
& Infer what action an utterance performs in context, including cases where the intended act differs from the literal form. \\

Discourse and Coherence 
& Maintain a coherent interpretation across turns by linking utterances via discourse relations and tracking what has been established, updated, or answered. \\

Social Pragmatics 
& Interpret and respond appropriately when meaning is shaped by social context such as roles, power, culture, and interactional norms. \\

Metaphor 
& Interpret figurative language by mapping a non-literal description to the situation to recover the intended evaluation or action guidance, rather than a literal reading. \\
\bottomrule
\end{tabular}
\caption{
Pragmatic sections used for self-generated QA data.
Each section description is inserted into the QA generation prompt to guide the model toward examples requiring the corresponding type of pragmatic interpretation.
}
\label{tab:pragmatic_sections}
\end{table*}

\subsection{QA Generation Prompt}
\label{append:qa_generation_prompt}

To construct the primary Pragmatic QA data, we prompt the model to generate exactly one QA item for a given domain and pragmatic section. 
Each generated item contains a concrete context, a question, and a short answer. 
The prompt explicitly requires that the question be impossible to answer without pragmatic interpretation.
We use a small pool of manually inspected few-shot examples to stabilize the format.
Generated items are rejected if they cannot be parsed, have missing fields, or duplicate an earlier item under a normalized string signature.

\noindent
\begin{appendixpromptbox}{QA Generation Prompt}
\footnotesize
\ttfamily

\textbf{System.}

You are a pragmatic QA data generator.

Generate exactly one item for the target section below.

Target section: \texttt{\{SECTION\_NAME\}}

Section description: \texttt{\{SECTION\_DESCRIPTION\}}

You may reason before the final output if needed.

The final output must start with a line containing exactly \texttt{FINAL\_QA:}

After \texttt{FINAL\_QA:}, return plain text only with this exact format:

Question must be impossible to answer without pragmatic interpretation.
Answer must be short and clear.

\texttt{FINAL\_QA:}

\texttt{Content: ...}

\texttt{Question: ...}

\texttt{Answer: ...}

No JSON and no extra commentary after \texttt{FINAL\_QA:}.

\vspace{0.8em}
\textbf{User.}

Domain keyword: \texttt{\{DOMAIN\}}

Item: \texttt{\{ITEM\_INDEX\}/\{ITEMS\_PER\_DOMAIN\}}

Field definitions:

- \texttt{content}: a concrete scenario/context only, with enough detail for pragmatic inference.

- \texttt{question}: one explicit question about the scenario.

- \texttt{answer}: a minimal, short, concise, single correct answer to the question without any explanation.

Section examples (style reference):

\texttt{\{FEW\_SHOT\_EXAMPLES\}}

Output plain text only with this exact format:

\texttt{FINAL\_QA:}

\texttt{Content: ...}

\texttt{Question: ...}

\texttt{Answer: ...}

No JSON, no markdown, no extra commentary after \texttt{FINAL\_QA:}.

\end{appendixpromptbox}

\subsection{Automatic Audit and Filtering}
\label{append:audit_filtering}

We audit each generated Pragmatic QA instance with a binary quality-judgment prompt. 
The auditor is given calibration examples from the same section, followed by the generated item to judge. 
An item is retained only if it is well-formed, unambiguous, answerable, and requires the intended type of pragmatic interpretation.

\noindent
\begin{center}
\begin{appendixpromptbox}{Automatic Audit and Filtering Prompt}
\footnotesize
\ttfamily

\textbf{System.}

You are a dataset auditor for Pragmatic QA. 
Judge whether the item is high-quality for its SECTION. 
Answer yes for high-quality, no for low-quality.

\vspace{0.6em}
\textbf{User.}

Is this pragmatic QA example high-quality?

\vspace{0.4em}
\textbf{Criteria.}

Pragmatic dependency: the gold answer requires pragmatic interpretation, not just literal reading.

Question correctness: the question itself is not incorrect or ambiguous.

Gold answer: the gold answer is correct and uniquely best-supported by the given context.

\vspace{0.4em}
Few-shot examples for this same SECTION: \emph{same-section accepted and rejected calibration examples}.

\vspace{0.4em}
Example to judge:

SECTION: \texttt{\{SECTION\}};

SCENARIO: \texttt{\{CONTENT\}};

QUESTION: \texttt{\{QUESTION\}};

GOLD ANSWER: \texttt{\{GOLD\_ANSWER\}}.

\vspace{0.4em}
Answer just yes or no with no other output. Final answer:

\end{appendixpromptbox}
\end{center}

We apply the audit judge with the first-token margin $m$ (\S\ref{sec:data-generation-filtering}) and discard the bottom $50\%$ of generated items by $m$, while preserving balance across pragmatic sections.

\section{Counterfactual Answer Generation and Training}
\label{append:training}

We train models in two sequential stages: supervised fine-tuning (SFT) followed by GRPO. 
Before SFT, we generate target answers for the filtered primary problem data. 
The counterfactual condition uses a privileged counterfactual reasoning script during answer generation, while the student model is later trained without this script in the input. 
The non-counterfactual variant uses the same pipeline but replaces the counterfactual script with a lighter pragmatic QA prompt.
\subsection{Answer Generation Prompts}
\label{append:answer_generation_prompts}

For the counterfactual condition, we use a pragmatic QA prompt that explicitly instructs the model to interpret the observed utterance as a communicative choice among plausible alternatives. 
The model is asked to reason about what the speaker could have said under a different intention and to use this contrast to infer the intended meaning.

\noindent
\begin{center}
\begin{appendixpromptbox}{Counterfactual Answer Generation Prompt}
\footnotesize
\ttfamily

\textbf{System.}

You are solving a pragmatic question-answering task.

Your goal is to choose the best answer by inferring the speaker's intended meaning in context.

Do not rely on the literal meaning of the utterance alone. 
Instead, interpret the utterance as a communicative choice made by a roughly rational and informative speaker. 
Think about why this speaker chose this utterance, in this context, given what the speaker likely knows.

When answering a question, use the following reasoning principles:

1. Identify the literal meaning of the utterance.

2. Use the context and shared background to determine what the speaker is likely trying to communicate.

3. Consider why the speaker chose this utterance instead of other plausible alternatives.

4. Assume the speaker is trying to provide relevant information in context, but may not say more than is needed.

5. Use the speaker's likely knowledge and the shared context to infer what the listener is expected to understand.

6. Choose the answer that best explains the utterance as a rational, informative, and contextually relevant choice.

When possible, justify your interpretation contrastively until you reach one clear interpretation:

- state one more direct, stronger, or more literal alternative the speaker could have said,

- explain what that alternative would have implied,

- then explain why the actual utterance suggests a different intended meaning.

Reason in the form:

``If the speaker had intended X, they would likely have said Y. Because they said Z instead, the intended meaning is more likely W.''

Guidelines:

- Prefer the answer that matches the speaker's intended meaning, not just the surface wording.

- Use context, shared background, speaker knowledge, and plausible alternatives to interpret the utterance.

- Prefer interpretations that best explain the speaker's choice of wording.

- Do not infer more than the context, shared background, and the speaker's choice support.

\end{appendixpromptbox}
\end{center}
For the non-counterfactual baseline, we use a lighter pragmatic QA prompt that asks the model to consider pragmatic meaning, but does not instruct it to explicitly contrast the observed utterance with alternative utterances. 
This provides a comparison between the counterfactual reasoning script and generic pragmatic prompting.
\noindent
\begin{center}
\begin{appendixpromptbox}{Non-counterfactual Answer Generation Prompt}
\footnotesize
\ttfamily

\textbf{System.}

You are solving a pragmatic
question-answering task.

Your goal is to choose the best answer by
inferring the intended meaning of the utterance
in the given context.

Do not rely on the literal meaning of the
utterance alone. Instead, use the surrounding
situation, discourse context, social setting,
and background information provided in the
problem to determine what the utterance
communicates.

Focus on how the utterance functions in the
specific scenario. Pay attention to what is
explicitly stated, what is implied, what is
presupposed, and what the context makes
salient.

When answering a question, use the following
reasoning principles:

1. Identify the literal meaning of the
utterance.

2. Identify the relevant contextual cues:
who is involved, what is happening, what has
just been said or done, and what information is
salient in the situation.

3. Determine the communicative function of the
utterance, such as informing, requesting,
refusing, warning, correcting, complaining,
reassuring, joking, teasing, agreeing,
disagreeing, implying, or expressing an
attitude.

4. Use the context and shared background in the
problem to infer what meaning goes beyond the
surface wording.

5. Pay attention to pragmatic phenomena such as
implicature, presupposition, deixis, indirect
speech acts, politeness, irony, sarcasm, humor,
metaphor, metonymy, discourse coherence, and
social expectations when they are relevant.

6. Distinguish between what is directly said
and what is pragmatically implied by the
scenario.

7. Choose the answer that best captures the
intended meaning while avoiding unsupported
assumptions or over-interpretation.

When possible, justify your interpretation by
grounding it in the provided context:

- state what the utterance literally says,
- identify the contextual cues that matter,
- explain how those cues shape the pragmatic
meaning,
- choose the answer that best matches the
intended meaning in context.

Guidelines:

- Prefer the answer that matches the intended
meaning, not just the surface wording.

- Use context, discourse coherence, shared
background, social norms, and pragmatic cues
to interpret the utterance.

- Prefer interpretations that are directly
supported by the scenario.

- Do not infer more than the context and
utterance support.

- Do not choose an answer merely because it is
possible; choose the answer that is most
coherent in context.

- Provide a concise reasoning trace followed
by the final answer.

\end{appendixpromptbox}
\end{center}

\subsection{Correctness Judge Prompt}
\label{append:judge_prompt}

\noindent
\begin{center}
\begin{appendixpromptbox}{Judge Prompt}
\footnotesize
\ttfamily

\textbf{System.}

You are a strict QA evaluator. 
Respond with exactly one word: either `yes' or `no'. 
Do not emit any other text, punctuation, or explanation.

\vspace{0.6em}
\textbf{User.}

Task: QA answer grading.

\vspace{0.4em}
CONTEXT: \texttt{\{CONTENT\}};

QUESTION: \texttt{\{QUESTION\}};

REFERENCE ANSWER: \texttt{\{REFERENCE\}};

CANDIDATE ANSWER: \texttt{\{CANDIDATE\}}.

\vspace{0.4em}
Is the candidate answer semantically correct given the context, question, and reference answer? 
Answer with a single word: yes or no.

\end{appendixpromptbox}
\end{center}

In all cases, the judge is the untuned instruct backbone matched in size to the policy under training, keeping the pipeline self-contained and free of external distillation.

We read the first-token log-probabilities for \text{yes} and \text{no} from the judge, convert them to probabilities, and compute the margin
\[
m(x, a) = p(\texttt{yes} \mid x, a) - p(\texttt{no} \mid x, a),
\]
where $x$ is the problem (context, question, and reference answer) and $a$ is the candidate answer. We accept a candidate if $m(x, a) > 0.8$. The choice of threshold is justified by the human-agreement study in \S\ref{append:judge_margin_calibration}.

\subsection{Margin Calibration}
\label{append:judge_margin_calibration}

To verify that the judge margin reflects meaningful correctness confidence, we conduct a small human-agreement calibration study. 
We sample 100 examples, and two authors independently assign one of three labels: \emph{correct}, \emph{incorrect}, or \emph{skip}, based on whether the candidate answer is semantically correct given the context, question, and reference answer.
We adopt a conservative aggregation rule: if either annotator assigns \emph{skip}, the example is mapped to \emph{incorrect}, regardless of the other annotation.
For examples receiving no \emph{skip} label, agreements are retained directly, while disagreements between \emph{correct} and \emph{incorrect} are discussed by the two annotators until consensus is reached.
The resulting consensus labels serve as the human gold labels for threshold calibration.
We then compare the margin-thresholded judge decisions against these labels.

\begin{table}[h]
\centering
\small

\begin{tabular}{lcccc}
\toprule
\textbf{Rule} & \textbf{Precision} & \textbf{Recall} & \textbf{F1} & \textbf{Accuracy} \\
\midrule
$m > 0.5$  & 0.741 & 0.833 & 0.784 & 0.780 \\
$m > 0.6$  & 0.741 & 0.833 & 0.784 & 0.780 \\
$m > 0.7$  & 0.769 & 0.833 & 0.800 & 0.800 \\
$m > 0.8$  & 0.780 & 0.812 & 0.796 & 0.800 \\
$m > 0.9$  & 0.771 & 0.771 & 0.771 & 0.780 \\
$m > 0.99$ & 0.767 & 0.688 & 0.725 & 0.750 \\
\bottomrule
\end{tabular}
\caption{
Agreement between self-judge margin thresholds and human labels on the manually reviewed calibration subset of 100 examples.
Examples that either annotator marked as \emph{skip} are mapped to \emph{incorrect} in the gold label rather than excluded, so all 100 examples are retained.
Precision, recall, F1, and accuracy are computed by treating the margin-thresholded judge decision as the prediction and the human label as gold.
}
\label{tab:judge_margin_calibration}
\end{table}

The calibration results show that the judge margin is informative: increasing the threshold generally makes the judge more conservative, reducing recall while maintaining comparable precision. 
The best human-agreement F1 is obtained at $m>0.7$ (0.800), while $m>0.8$ remains a high-agreement operating point with precision 0.780, recall 0.812, F1 0.796, and accuracy 0.800. 
Very strict thresholds such as $m>0.99$ substantially reduce recall, suggesting that overly conservative filtering discards many human-acceptable responses. 
We adopt $\tau = 0.8$ as a conservative midpoint that preserves judge precision. 
The threshold is fixed before running downstream experiments and is not tuned against pragmatic benchmark performance. 
The same value is used for all four backbones: Qwen3-8B, Qwen3-14B, Gemma-4-E4B, and GPT-OSS-20B.

\subsection{Supervised Fine-Tuning}
\label{append:sft_details}

\paragraph{SFT Pregeneration.}
Let $x$ denote an input instance from the filtered primary problem data, and let
$s \in \mathcal{S}$ denote its pragmatic section label. 
We write $d(s)$ for the natural-language description of section $s$, and
$p_{\mathrm{cf}}$ for the counterfactual pragmatic reasoning script used during
response pre-generation. 
Let $\pi_\theta$ denote the base policy used to generate candidate SFT targets.
For each retained problem, we construct an augmented teacher-side prompt
\[
\tilde{x} = \mathrm{Aug}(x,s) = [p_{\mathrm{cf}};\, d(s);\, x],
\]
which exposes the teacher to both the section description and an explicit
counterfactual reasoning scaffold. Given $\tilde{x}$, the model samples a response
\[
y \sim \pi_\theta(\cdot \mid \tilde{x}),
\]
where $y=(r,a)$ consists of a reasoning trace $r$ and a final answer $a$.
We then apply a binary self-judge with the margin method (\S\ref{append:judge_prompt}).
\[
J(\tilde{x}, a) \in \{0,1\},
\]
which returns 1 only if the response is judged pragmatically correct with
respect to the context, question, and reference answer. 
This yields the accepted set
\[
\mathcal{D}_{\mathrm{accept}}
=
\{(x,s,y) \mid J(\tilde{x}, a)=1\}.
\]

The final SFT dataset removes the privileged augmentation from the student
input. 
Although the target response $y$ is generated under $\tilde{x}$, the student is
trained only on the original problem $x$ paired with the accepted output:
\[
\mathcal{D}_{\mathrm{SFT}}
=
\{(x,y) \mid (x,s,y) \in \mathcal{D}_{\mathrm{accept}}\}.
\]
This asymmetry between teacher-side generation and student-side training is
central to our design: the counterfactual reasoning script is used to construct
high-quality reasoning traces, but the student must learn to produce such traces
without seeing the script at inference time. 
We train with the standard causal language modeling objective on
$\mathcal{D}_{\mathrm{SFT}}$, masking prompt tokens and applying loss only to the
assistant response:
\[
\mathcal{L}_{\mathrm{SFT}}
=
-\sum_{(x,y)\in \mathcal{D}_{\mathrm{SFT}}}
\sum_{t=1}^{|y|}
\log \pi_{\theta}(y_t \mid x, y_{<t}).
\]

\paragraph{SFT Hyperparameters.}
We use full-parameter fine-tuning with maximum sequence length 8192, bfloat16 precision, AdamW, cosine learning-rate schedule, learning rate $5\times10^{-7}$, two epochs, per-device batch size 1, gradient accumulation 4, warmup ratio 0.03, gradient clipping 1.0, and gradient checkpointing. 
Distributed runs use FSDP full-shard training with decoder-layer auto-wrapping and full-state-dict checkpointing.

\subsection{Dataset Size and Training Budget}
\label{append:dataset_size}

We first sample 1,000 domain seeds and pair each seed with each of the six pragmatic sections, yielding 6,000 seed--section generation prompts. 
All 6,000 generations are parsed into valid short-answer QA items. 
The self-filtering stage retains 3,000 primary problem instances, corresponding to 500 examples per pragmatic section. 
These 3,000 filtered problems are used as the GRPO training prompts.

For SFT, we generate target responses for the filtered primary problems using the counterfactual reasoning script and then apply the answer-quality judge. 
During Qwen3-14B training, this produces 2,816 accepted SFT targets, of which 2,759 are used for SFT training and 57 are held out as a synthetic validation split.

\begin{table}[h]
\centering
\small
\setlength{\tabcolsep}{6pt}
\begin{tabular}{lr}
\toprule
\textbf{Stage} & \textbf{Count} \\
\midrule
Domain seeds & 1000 \\
Seed--section generation prompts & 6000 \\
Parsed valid QA items & 6000 \\
Retained after self-filtering & 3000 \\
Accepted SFT targets & 2816 \\
SFT train split & 2759 \\
SFT validation split & 57 \\
GRPO training prompts & 3000 \\
\bottomrule
\end{tabular}
\caption{
Aggregate data counts after answer generation and filtering.
The accepted SFT targets are generated from the filtered primary problem data.
}
\label{tab:sft_data_counts}
\end{table}

\subsection{GRPO}
\label{append:grpo_details}

We initialize the policy from the SFT checkpoint $\pi_{\mathrm{SFT}}$ and optimize it with GRPO~\citep{shao_deepseekmath_2024} on the primary problem data. Below we first specify the algorithmic setup (reward, optimization objective) and then the implementation details (hyperparameters, infrastructure).

\paragraph{Easy-Prompt Filtering.}
Following DAPO~\citep{yu_dapo_2026}, we apply an additional difficulty-based pass over the primary problem data before GRPO training. For each prompt $x$, we sample $G=8$ rollouts from the SFT checkpoint $\pi_{\mathrm{SFT}}$ and score each rollout with the same correctness judge and margin $m$ (\S\ref{append:judge_prompt}) used during GRPO (threshold $\tau=0.8$). A prompt is marked \emph{easy} if every one of its $G$ rollouts passes the judge, and is discarded. Such prompts yield zero-variance advantage estimates under group normalization and therefore contribute no gradient signal to the policy update. Filtering thus concentrates training on prompts at the frontier of $\pi_{\mathrm{SFT}}$'s capability.

\paragraph{Reward Design.}
\emph{Format reward.} $R_{\mathrm{fmt}}(y) \in \{0,\,0.5,\,1\}$ is a dense shaping signal that stabilizes early training by encouraging the model to maintain the structured output it acquired during SFT. A response receives $0.5$ for containing exactly one well-formed pair of \texttt{<think>}$\ldots$\texttt{</think>} tags, and an additional $0.5$ for a valid \texttt{\textbackslash boxed\{\}} answer in the post-thinking segment.

For each training prompt $x$, the policy $\pi_\theta$ samples a group of $G=8$ candidate responses $\{y_1, \ldots, y_G\}$, where each $y_i$ consists of a reasoning trace and a boxed final answer $a_i$. Each response is scored by a composite reward
\begin{equation*}
R(x, y_i) \;=\; w_{\mathrm{fmt}}\, R_{\mathrm{fmt}}(y_i) \;+\; w_{\mathrm{ans}}\, R_{\mathrm{ans}}(x, y_i),
\end{equation*}
combining a format-compliance term with a pragmatic-correctness term. 

\emph{Correctness reward.} $R_{\mathrm{ans}}(x, y) \in \{0, 1\}$ reuses the judge margin $m(x, a)$ from \S\ref{append:judge_prompt}. If \texttt{\textbackslash boxed\{\}} extraction fails, we set $R_{\mathrm{ans}}(x, y) = 0$ without querying the judge. Otherwise we extract the candidate answer $a$ from the rollout and assign $R_{\mathrm{ans}}(x, y) = 1$ if $m(x, a) > \tau$, with $\tau = 0.8$ fixed by the calibration in Appendix~\ref{append:judge_margin_calibration}.

\paragraph{Reward Scaling.}
\label{append:reward_scale}
We set $w_{\mathrm{fmt}} = 1$ and $w_{\mathrm{ans}} = 2$ so that correctness strictly dominates format. This asymmetric scaling prevents the policy from trading off pragmatic correctness for the denser, near-saturated format signal during early optimization. We use the smallest integer weighting that establishes this strict ordering rather than tuning the ratio against benchmark performance. 

\paragraph{Optimization Objective.}
Given the per-response rewards, GRPO computes group-normalized advantages
\begin{equation*}
\hat{A}_i = \frac{R(x, y_i) - \mathrm{mean}\!\left(\{R(x, y_j)\}_{j=1}^G\right)}{\mathrm{std}\!\left(\{R(x, y_j)\}_{j=1}^G\right)},
\end{equation*}
and updates the policy with a clipped PPO-style surrogate objective without a learned value function. Because the learning signal comes from \emph{relative} quality differences within each group, training is robust to the absolute scale of the rewards. 

\paragraph{GRPO Hyperparameters.}
We use full-parameter optimization with bfloat16 precision, AdamW, and a cosine learning-rate schedule with peak learning rate $4\times10^{-6}$ and warmup ratio $0.1$. The train batch size and PPO mini-batch size are both $128$, with per-GPU micro-batch size $2$. Maximum prompt and response lengths are $512$ and $1{,}536$ tokens, respectively; overlong prompts are filtered out and the remainder are left-truncated. Following \citet{srivastava2025debate, xin2025deepseekprover, shao_deepseekmath_2024, he-etal-2025-rewarding}, we regularize the policy toward $\pi_{\mathrm{SFT}}$ with a low-variance KL loss applied directly to the objective with coefficient $0.02$, which discourages drift from the counterfactual reasoning behaviors acquired during SFT. 
We train for $4$ epochs over the primary problem data.

\paragraph{GRPO Infrastructure.}
Each GRPO run uses $5$ NVIDIA H200 GPUs: $4$ GPUs host the policy and serve rollouts in-process via vLLM (one GPU per node, FSDP2 full-shard across the $4$ nodes), and the remaining GPU runs a separate vLLM endpoint hosting the frozen instruct model as the judge. Rollouts are sampled at temperature $1.0$, top-$p$ $1.0$, and top-$k$ disabled, with $G=8$ samples per prompt. End-to-end wall-clock for the $4$-epoch run on the primary problem data is approximately \textbf{2.5} hours on this configuration.

\subsection{GRPO Checkpoint Selection}
\label{append:ckpt_selection}

To avoid last-checkpoint or hand-picked bias, we report the checkpoint chosen by a fixed selection protocol. GRPO saves a checkpoint at every optimizer step and selects among these checkpoints as follows.

\paragraph{Held-out selection set.}
We construct the selection set from a fresh round of the self-generation pipeline described in \cref{append:data_construction}, run independently of the round used to produce the GRPO training data. 
From this fresh pool we draw a seeded, stratified sample of $100$ examples per pragmatic section. 
To guarantee disjointness from training, any selection row whose $(\text{context}, \text{question})$ pair appears in the GRPO training data is discarded. 
The selection set is fixed with seed $42$ across all checkpoints of all runs of a given model, so all checkpoints of a model are graded on exactly the same rows.

\paragraph{Scoring.}
Each checkpoint is scored on the selection set using the same correctness judge and margin $m(x, a)$ (\S\ref{append:judge_prompt}) that define the GRPO training reward, with a row counted correct if $m(x, a) > 0.8$. This means selection accuracy is the GRPO training reward itself, computed on held-out data, rather than a separate evaluation metric. For decoding, we generate one rollout per row with temperature $0$ and \texttt{max\_tokens}~$=2048$, scoring each row independently.

\paragraph{Selection Rule.}
We pick the checkpoint with the highest selection accuracy. 
Exact ties are broken in favor of the later step, on the principle that the later step has absorbed strictly more of the training signal and is therefore the more conservative choice to promote.

\paragraph{Independence from Test Benchmarks.}
The selection set consists of self-generated pragmatic QA filtered by the same base-model judge used throughout our pipeline (matched in size to the trained model), and shares no items with \pragmega, \ludwig, \metoqa, or \altprag. Selection therefore cannot leak signal from these benchmarks, so the test numbers in Table~\ref{tab:qwen3_pragmatic_bold} measure generalization beyond the selection pool.

\section{Benchmarks and Results}
\label{append:benchmarks and results}
\subsection{Benchmark Examples}
\cref{tab:benchmark_examples} shows two of each benchmark example. 

\label{append:benchmark_examples}

\begin{table*}[t]
\centering
\scriptsize
\setlength{\tabcolsep}{4pt}
\renewcommand{\arraystretch}{1.18}
\begin{tabularx}{\textwidth}{p{0.13\textwidth}p{0.58\textwidth}X}
\toprule
\textbf{Benchmark} & \textbf{Example input} & \textbf{Expected output} \\
\hline

\rowcolor{gray!8}

\pragmega
&
Scenario: Mary was asked about the town that she has just moved to. Mary responded:
``This town is a chimney.'' What does Mary mean?

Options:
1) The people living in this town are very welcoming.
2) The town is a chimney.
3) All houses in this town have chimneys.
4) Mary found a job at a company installing chimneys.
5) The town is not one of the cleanest ones.
&
5. The intended meaning is non-literal: Mary implies that the town is dirty or polluted. \\

\pragmega
&
Scenario: Paul has to go to an interview and he is running late. While cleaning his shoes, he says to his wife Jane:
``I want to wear that blue shirt, but it is very creased.'' What might he be trying to convey?

Options:
1) He wants his wife to iron his shirt.
2) He is unhappy that his shirt is wrinkled.
3) He got this shirt on sale.
4) He got his shoes on sale.
&
1. Paul is making an indirect request for Jane to iron the shirt. \\

\hline

\rowcolor{gray!8}
\ludwig
&
Question: Is Marci grumpy?

Response: ``He's as gentle as a lamb.''
&
No. \\

\ludwig
&
Question: You... live here?

Response: ``Long time. Long, long time.''
&
Yes. \\

\hline

\rowcolor{gray!8}
\metoqa
&
Context: She is attracted to blue jacket.

Question: What does ``blue jacket'' refer to?

Options:
1) Colour
2) Jacket
3) Sailor
4) Sea
&
3. Sailor. \\

\metoqa
&
Context: His lovely voice caught my ear.

Question: What does the sentence refer to?

Options:
1) Giving attention
2) Noise
3) Whispering to the person
4) None
&
1. Giving attention. \\

\hline

\rowcolor{gray!8}
\altprag
&
Context: A teacher is speaking with Sam about his assignments and commitments, noting that Sam might be overwhelmed.

Root: ``You have too much on your plate.''

Candidate 1: ``I appreciate your concern, but I think I can manage everything just fine.''

Candidate 2: ``It's just how I like it; wouldn't have it any other way.''
&
Candidate 1 directly reassures the teacher while politely asserting confidence. It is preferred because it acknowledges the teacher's concern rather than dismissing it. \\

\altprag
&
Context: Melissa is reflecting on the new sales manager who just started a few days ago: ``She is a babe in the woods.''

Root: ``She is a babe in the woods.''

Candidate 1: ``Wow, I hope she's ready for the steep learning curve!''

Candidate 2: ``Yeah, but honestly, who isn't at that level when starting fresh?''
&
Candidate 1 expresses concern about the manager's readiness and emphasizes the difficulty of the role, rather than normalizing her inexperience. \\

\bottomrule
\end{tabularx}
\caption{
Representative examples from the four pragmatic evaluation benchmarks used in our experiments.
The examples illustrate the different forms of pragmatic interpretation tested by each benchmark:
multiple-choice pragmatic QA in \pragmega, binary implicature resolution in \ludwig,
metonymic reference resolution in \metoqa, and open-ended implied-meaning recovery in \altprag.
}
\label{tab:benchmark_examples}
\end{table*}

\subsection{Exploratory Evaluation on Non-Cooperative Pragmatics}
\label{app:sda_eval}

To examine whether counterfactual reasoning transfers to non-cooperative and adversarial settings, we evaluate \method{} using the Strategic Dialogue Assessment (SDA) framework introduced by \citet{zheng-etal-2026-strategic}. SDA evaluates courtroom cross-examinations as strategic exchanges, measuring whether a model can track how each response affects the speaker's position in the dialogue. \citet{zheng-etal-2026-strategic} find that LLMs can rely on surface-level discourse cues when judging adversarial dialogue, sometimes treating damage control strategies such as hedging or deflection as neutral or positive rather than recognizing them as attempts to mitigate a harmful commitment. This setting therefore provides a complementary test of whether \method{} helps models reason beyond the cooperative surface form of an utterance.

We focus on three primary SDA metrics: \textbf{BaT} (Benefit at Turn), which measures alignment with human judgments of strategically beneficial moves; \textbf{PaT} (Penalty at Turn), which measures alignment with human judgments of strategically detrimental moves; and \textbf{NRBaT} (Normalized Relative Benefit at Turn), which captures the cumulative balance between benefits and penalties over the dialogue. 

\begin{table}[h]
\centering
\footnotesize
\setlength{\tabcolsep}{3pt}
\newcommand{\std}[2]{#1{\scriptsize$\pm$#2}}
\newcommand{\beststd}[2]{\textbf{#1{\scriptsize$\pm$#2}}}
\begin{tabular}{@{}lccc@{}}
\toprule
\textbf{Model} & \textbf{BaT} & \textbf{PaT} & \textbf{NRBaT} \\
\midrule
\multicolumn{4}{c}{\textit{Qwen3-8B}} \\
\midrule
Instruct 
& \beststd{0.132}{0.036} 
& \std{-0.001}{0.033} 
& \std{0.149}{0.058} \\
\textsc{PragReST} 
& \std{0.051}{0.038} 
& \beststd{0.013}{0.033} 
& \beststd{0.179}{0.034} \\
\midrule
\multicolumn{4}{c}{\textit{Qwen3-14B}} \\
\midrule
Instruct 
& \std{0.112}{0.023} 
& \std{0.047}{0.061} 
& \beststd{0.075}{0.103} \\
\textsc{PragReST} 
& \beststd{0.138}{0.054} 
& \beststd{0.062}{0.074} 
& \std{0.073}{0.119} \\
\bottomrule
\end{tabular}
\caption{Performance on the SDA framework. Values report mean Spearman's $\rho$ correlations with human judgments across five seeds (using temperature sampling at 0.6), with standard deviations shown after $\pm$. Higher values indicate stronger alignment with human judgments.}
\label{tab:sda_results}
\end{table}

\paragraph{Quantitative Results.} As shown in Table~\ref{tab:sda_results}, the most consistent change appears on \textbf{PaT}. \method{} increases PaT for both Qwen3-8B (from $-0.001$ to $0.013$) and Qwen3-14B (from $0.047$ to $0.062$). These gains are modest relative to variation across seeds, so we interpret them cautiously. Still, the consistent direction of the change suggests that counterfactual training may improve the model's ability to recognize when a response imposes a strategic cost on the speaker, rather than treating locally cooperative answers as neutral or beneficial.

The remaining SDA metrics show a more mixed pattern. For Qwen3-8B, \method{} improves NRBaT from $0.149$ to $0.179$, but decreases BaT from $0.132$ to $0.051$. For Qwen3-14B, \method{} improves BaT and PaT, while NRBaT remains essentially unchanged, moving from $0.075$ to $0.073$. We therefore interpret the SDA results as evidence for a targeted improvement in recognizing strategic penalties, rather than a uniform improvement across all dimensions of adversarial dialogue assessment.

\paragraph{Qualitative analysis.} To better understand the PaT gains, we inspect turns where \method{} agrees with the human penalty judgment but the instruct backbone does not. Most recovered cases involve a change in how the model interprets the witness's answer: \method{} is more likely to recognize that the witness has conceded information that helps the opposing side. In SDA terms, this means identifying a response as strategically harmful even when it is locally clear, truthful, and relevant. Thus, the PaT gains suggest that \method{} is not simply rewarding answers for being clear or responsive. Instead, it more often recognizes when an apparently cooperative answer gives the opposing side useful information.

For example, when a witness is asked whether the defendant ``voluntarily spoke with you in a tape-recorded interview without the presence of counsel'' and answers ``Yes,'' the instruct model recognizes that the response is clear and responsive, but still treats it as beneficial to the witness's side. \method{} instead identifies the strategic implication of the same answer: by confirming the questioner's premise, the witness gives the opposing side the concession it is seeking. Similar patterns appear when a witness gives a precise damaging answer (``Seven'' abrasions), confirms a document detail (``Yes, it is''), concedes a contamination pathway (``it is likely''), or admits a lack of licensed qualification. In each case, \method{} treats the utterance not merely as a cooperative answer, but as a commitment whose strategic value depends on the question under discussion.

This behavior suggests that the model interprets each utterance in relation to the adversarial context: it considers what the answer allows the questioner to infer and whether that inference advances the opposing side's case. In the recovered cases, \method{} often reasons over alternatives implicitly, recognizing that a direct answer rather than a hedge, a concession rather than a denial, or a clarification that still preserves a damaging inference can change which side the utterance benefits. This supports the quantitative pattern in Table~\ref{tab:sda_results}: the most consistent gains appear in PaT, where success depends on recognizing when a response creates a strategic cost for the speaker.

\section{Details of the Counterfactual Error Analysis}
\label{append:cf_error_analysis}

This appendix describes the construction, validation, and use of the error taxonomy and counterfactual-reasoning scores used in \cref{sec:error_analysis}.

\subsection{Error Taxonomy}
\label{append:error_taxonomy}

\paragraph{Error-case Collection.}
We collect incorrect \pragmega predictions from each evaluated model. 
For every incorrect example, we retain the original prompt, answer options, gold answer, model prediction, and model reasoning trace. 
These traces are used only for diagnostic analysis, not for computing task accuracy.

\paragraph{Inducing the Error Taxonomy.}
\label{append:error_taxo}
We induce the error taxonomy in a bottom-up manner. 
Instead of manually specifying categories before inspecting the data, we prompt a language-model annotator to read batches of incorrect examples and propose recurring failure modes. 
The annotator is instructed to focus on the underlying pragmatic reasoning failure rather than superficial lexical differences. 
After inspecting the proposed categories, we merge near-duplicates and remove categories that are too broad, too rare, or outside the scope of pragmatic reasoning. 
We also remove categories that merely indicate that \method{} introduces our target behavior, since the goal is to characterize model errors rather than reward-specific style differences.

The final taxonomy contains the following non-exclusive tags.

\begin{itemize}
    \item \textbf{Literal / surface bias}: the model anchors on literal wording or shallow semantic compatibility when the context requires a non-literal pragmatic interpretation.
    \item \textbf{Missed communicative intent}: the model fails to recover the speaker or listener's pragmatic goal, such as politeness, avoidance, deception, indirect request, complaint, or social positioning.
    \item \textbf{Unsupported or overextended inference}: the model over-reasons from weak cues, invents assumptions not licensed by the prompt, or post-hoc rationalizes an incorrect answer.
    \item \textbf{Coherence bridge error}: the model misjudges whether an implicit causal, temporal, or discourse bridge between events is warranted.
    \item \textbf{Figurative or humor mapping error}: the model fails to map figurative language, jokes, punchlines, or humorous incongruity to the intended interpretation.
\end{itemize}

Tags are multi-label: a single error may receive more than one tag if multiple failure modes are present.

\subsection{Annotation Protocol and Human Validation}
\label{append:error_annotation_validation}

\paragraph{GPT Annotation Protocol.}
For full-scale annotation, we use a GPT-4.1-mini annotator. 
The annotator receives the prompt, gold answer, model prediction, and reasoning trace, along with the final taxonomy and short definitions of each error type. 
It is instructed to assign all applicable labels and to avoid assigning a pragmatic label when the failure is better explained by a concrete context or option misread. 
The model returns a structured label set for each incorrect example.

\paragraph{Human Validation.}
\label{append:human_valid}
To check whether GPT labels are reliable enough for diagnostic analysis, we run a blind annotation study. 
Three human annotators independently label the same subset of examples using the same taxonomy, without seeing the model identity. 
Two annotators were members of the project team, and one annotator was an external annotator who was not involved in developing the method or analyzing the model outputs. 
The annotators were given anonymized benchmark items, gold answers, model predictions, reasoning traces, and the fixed error-category definitions, and were asked to assign all applicable error labels. 
The annotation task involved only model-generated text and benchmark examples; we did not collect personal or sensitive information from annotators, and the labels were used only to estimate agreement with the automatic error annotations.
We compute pairwise agreement among humans and between each human and GPT.

\begin{table}[t]
\centering
\scriptsize
\setlength{\tabcolsep}{8pt}
\begin{tabular}{lcc}
\toprule
\textbf{Comparison} & \textbf{Agreement} & \textbf{Micro-$\kappa$} \\
\midrule
\multicolumn{3}{l}{\textit{Summary}} \\
Human--Human avg. & 83.8\% & 0.628 \\
Human--GPT avg.   & 82.6\% & 0.614 \\
\midrule
\multicolumn{3}{l}{\textit{Pairwise comparisons}} \\
Human A (external) vs. Human B & 80.0\% & 0.551 \\
Human A (external) vs. Human C & 87.7\% & 0.707 \\
Human B vs. Human C            & 83.8\% & 0.625 \\
\midrule
Human A (external) vs. GPT & 80.0\% & 0.557 \\
Human B vs. GPT            & 86.7\% & 0.712 \\
Human C vs. GPT            & 81.1\% & 0.572 \\
\bottomrule
\end{tabular}
\caption{
Agreement study for the induced error taxonomy.
All annotators labeled the same shared subset of errors under a blind setting without access to model identities.
Human A is an external annotator who was not involved in the project.
}
\label{tab:error_annotation_agreement}
\end{table}

The GPT annotator agrees with humans at approximately the same level as humans agree with one another. 
We therefore use GPT labels for the full analysis, but treat the resulting labels as a scalable diagnostic rather than as definitive ground truth.

\begin{table}[t]
\centering
\scriptsize
\setlength{\tabcolsep}{5pt}
\begin{tabular}{lcccc}
\toprule
\textbf{Annotator} & \textbf{Agreement} & \textbf{Micro $\kappa$} & \textbf{Micro $\phi$} & \textbf{Macro $\kappa$} \\
\midrule
GPT-4.1-mini & 82.60 & 0.614 & 0.619 & 0.497 \\
GPT-4.1 & 79.63 & 0.586 & 0.622 & 0.465 \\
GPT-5.5 & 77.24 & 0.548 & 0.594 & 0.424 \\
\bottomrule
\end{tabular}
\caption{
Agreement between human and LLM annotators for the induced pragmatic error taxonomy.
Each LLM row reports the average agreement between that LLM annotator and the three human annotators.
Agreement is computed over binary decisions for the five overlapping error categories.
}
\label{tab:error_taxonomy_agreement}
\end{table}

\subsection{Counterfactual Reasoning Score}
\label{append:cf_score}

For each reasoning trace, we compute a counterfactual-reasoning score using a GPT-based evaluator. 
The evaluator is instructed to judge only the reasoning trace, not whether the final answer is correct. 
It assigns five binary indicators corresponding to explicit counterfactual reasoning, alternative utterance or action, mismatch or contrast, speaker intent or pragmatic goal, and literal-versus-pragmatic contrast. 
The CF score is the sum of these indicators and ranges from 0 to 5. 
We use this score only for diagnostic analysis, not for training, filtering, checkpoint selection, or model evaluation.

\begin{figure*}[h]
    \centering
    \includegraphics[width=1\linewidth]{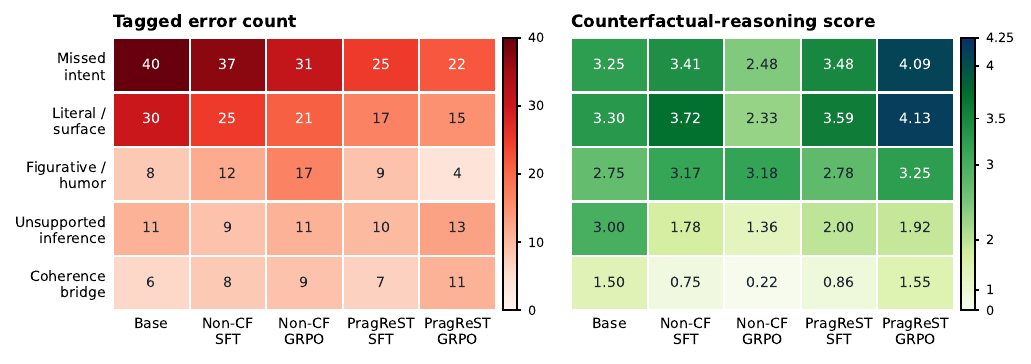}
    \caption{
    Full diagnostic breakdown of error reduction and counterfactual-reasoning scores across failure modes.
    Rows correspond to induced error categories, and columns are training conditions, with cells reporting the error-rate change and counterfactual-reasoning score statistics used in the main analysis.
    }
    \label{fig:cf_error_heatmap}
\end{figure*}

\noindent
\begin{center}
\begin{appendixpromptbox}{Counterfactual Reasoning Score Prompt}
\footnotesize
\ttfamily

\textbf{System.}

You are evaluating whether a model's reasoning trace uses counterfactual pragmatic reasoning.

Return only valid JSON. 
Do not judge whether the final answer is correct; judge only the reasoning trace.

\vspace{0.6em}
\textbf{User.}

Score the reasoning trace using the rubric below. 
Assign each field 0 or 1.

\vspace{0.4em}
1. Explicit counterfactual.

Assign 1 if the trace explicitly reasons about a counterfactual pragmatic or communicative alternative, e.g., 
``if the speaker meant X, they would/could have said Y,'' 
``had she intended X,'' or 
``if this were literal, we would expect \ldots''.
Do not count generic uncertainty, ordinary causal hypotheses, or narrative-coherence alternatives unless the trace compares actual behavior, wording, option choice, or interpretation against an alternative communicative/pragmatic possibility.

\vspace{0.4em}
2. Alternative utterance or action.

Assign 1 if the trace identifies an alternative wording, action, answer, or response that would be expected under a different intended meaning.

\vspace{0.4em}
3. Mismatch or contrast.

Assign 1 if the trace explicitly notes a mismatch or contrast between the literal/surface reading and contextual/pragmatic cues, or between an option and what the speaker or situation would normally imply.

\vspace{0.4em}
4. Speaker intent or pragmatic goal.

Assign 1 if the trace reasons about speaker/listener intent, social goal, politeness, deception, irony, indirectness, communicative purpose, or pragmatic meaning.

\vspace{0.4em}
5. Literal vs. pragmatic contrast.

Assign 1 if the trace explicitly contrasts literal meaning or face-value reading with intended, pragmatic, figurative, indirect, ironic, or non-literal meaning.

\vspace{0.6em}
Compute \texttt{cf\_score} as the sum of the five binary fields, from 0 to 5.

\vspace{0.6em}
\textbf{Question prompt:}

\texttt{\{FULL\_PROMPT\}}

\vspace{0.6em}
\textbf{Model reasoning trace:}

\texttt{\{REASONING\}}

\end{appendixpromptbox}
\end{center}

\section{Artifact Licenses}
All licenses are given in \cref{tab:licenses}.

\begin{table}[h]
\centering
\footnotesize
\setlength{\tabcolsep}{3pt}
\renewcommand{\arraystretch}{1.05}
\begin{tabularx}{\columnwidth}{@{}lX@{}}
\hline
\textbf{Artifact / Resource} & \textbf{License / Terms} \\
\hline
Social IQa 
& \href{https://creativecommons.org/licenses/by/4.0/}{CC BY 4.0} \\
AltPrag 
& \href{https://www.apache.org/licenses/LICENSE-2.0}{Apache License 2.0} \\
Ludwig 
& \href{https://creativecommons.org/licenses/by/4.0/}{CC BY 4.0} \\
PragMega 
& Not specified \\
MetoQA (PUB) 
& \href{https://opensource.org/license/mit/}{MIT License} \\
MMLU-Pro 
& \href{https://opensource.org/license/mit/}{MIT License} \\
MATH-500 
& \href{https://opensource.org/license/mit/}{MIT License} \\
AIME2025 
& \href{https://opensource.org/license/mit/}{MIT License} \\
TruthfulQA
& \href{https://www.apache.org/licenses/LICENSE-2.0}{Apache License 2.0} \\
Qwen3-8B 
& \href{https://www.apache.org/licenses/LICENSE-2.0}{Apache License 2.0} \\
Qwen3-14B 
& \href{https://www.apache.org/licenses/LICENSE-2.0}{Apache License 2.0} \\
Gemma-4-E4B
& \href{https://www.apache.org/licenses/LICENSE-2.0}{Apache License 2.0} \\
GPT-OSS-20B
& \href{https://www.apache.org/licenses/LICENSE-2.0}{Apache License 2.0} \\
OpenAI Models 
& \href{https://openai.com/policies/services-agreement/}{OpenAI Services Agreement} / \href{https://openai.com/policies/row-terms-of-use/}{Terms of Use} \\
\hline
\end{tabularx}
\caption{All datasets and models were used in accordance with their intended use.}
\label{tab:licenses}
\end{table}

\section{Note on AI Usage}
We used AI tools for grammar correction and code completion.

\end{document}